\documentclass[final]{cvpr}

\usepackage{times}
\usepackage{epsfig}
\usepackage{graphicx}
\usepackage{amsmath}
\usepackage{amssymb}

\usepackage{multicol}
\usepackage{booktabs}
\usepackage{multirow}
\usepackage{subfig}

\usepackage[pagebackref=true,breaklinks=true,colorlinks,bookmarks=false]{hyperref}

\begin{document}

\title{Vision-based Price Suggestion for Online Second-hand Items}

\author{Liang Han
\thanks{Liang Han is a PhD student at Department of Computer Science, Stony Brook University. He participated in this project as a research intern.}\\
Stony Brook University\\
{\tt\small liahan@cs.stonybrook.edu}
\and
Zhaozheng Yin
\thanks{Corresponding author. Zhaozheng Yin is an associate professor at Department of Computer Science and Department of Biomedical Informatics, Stony Brook University. He participated in this project during his sabbatical.}\\
Stony Brook University\\
{\tt\small zyin@cs.stonybrook.edu}

\and
Zhurong Xia\\
Alibaba Group\\
{\tt\small zhurong.xzr@alibaba-inc.com}

\and
Li Guo\\
Alibaba Group\\
{\tt\small guihe.gl@alibaba-inc.com}

\and
Mingqian Tang\\
Alibaba Group\\
{\tt\small mingqian.tmq@alibaba-inc.com}

\and
Rong Jin\\
Alibaba Group\\
{\tt\small jinrong.jr@alibaba-inc.com}
}

\maketitle


\begin{abstract}
Different from shopping in physical stores, where people have the opportunity to closely check a product (e.g., touching the surface of a T-shirt or smelling the scent of perfume) before making a purchase decision, online shoppers rely greatly on the uploaded product images to make any purchase decision. The decision-making is challenging when selling or purchasing second-hand items online since estimating the items' prices is not trivial. In this work, we present a vision-based price suggestion system for the online second-hand item shopping platform. The goal of vision-based price suggestion is to help sellers set effective prices for their second-hand listings with the images uploaded to the online platforms. 

First, we propose to better extract representative visual features from the images with the aid of some other image-based item information (e.g., category, brand). Then, we design a vision-based price suggestion module which takes the extracted visual features along with some statistical item features from the shopping platform as the inputs to determine whether an uploaded item image is qualified for price suggestion by a binary classification model, and provide price suggestions for items with qualified images by a regression model. According to two demands from the platform, two different objective functions are proposed to jointly optimize the classification model and the regression model. For better model training, we also propose a warm-up training strategy for the joint optimization. Extensive experiments on a large real-world dataset demonstrate the effectiveness of our vision-based price prediction system.
\end{abstract}

\section{Introduction}
\label{sec:intro}

Benefiting from the prevalence of smartphones, laptops and internet, the E-commerce has been on an exponential growth curve during the last couple of years. Traditionally, the E-commerce is mainly Business-to-Customer (B2C), and products sold on the online platforms are brand new. In recent years, inspired by the sharing economy, more and more people are trying to cash in their second-hand idle assets, which is heating up the Customer-to-Customer (C2C) second-hand trading. Moreover, a number of online platforms such as eBay, Letgo, Xianyu and Mercari, have emerged to help facilitate the online second-hand market.

Usually, the second-hand platforms do not make restrictions on how the sellers set prices for their listings. However, it is a great challenge for the sellers to set accurate or reasonable prices for their second-hand items. Compared with brand new products, most of the second-hand items listed on the platforms are unique, which means price references can be seldomly found, though the second-hand item transaction records can be saved. Accordingly, it can be a great help for sellers if the platforms can provide some price suggestions for their listings. Considering the online buyers rely heavily on the item images to make purchase decisions, we believe that there is a strong relationship between the item price and its image. In this work, we try to design a price prediction system, which can provide effective price suggestions for second-hand items listed on the online platforms with the uploaded images.

When trying to solve this real-world problem, we confront multiple challenges:

\begin{figure}
\includegraphics[width=\linewidth]{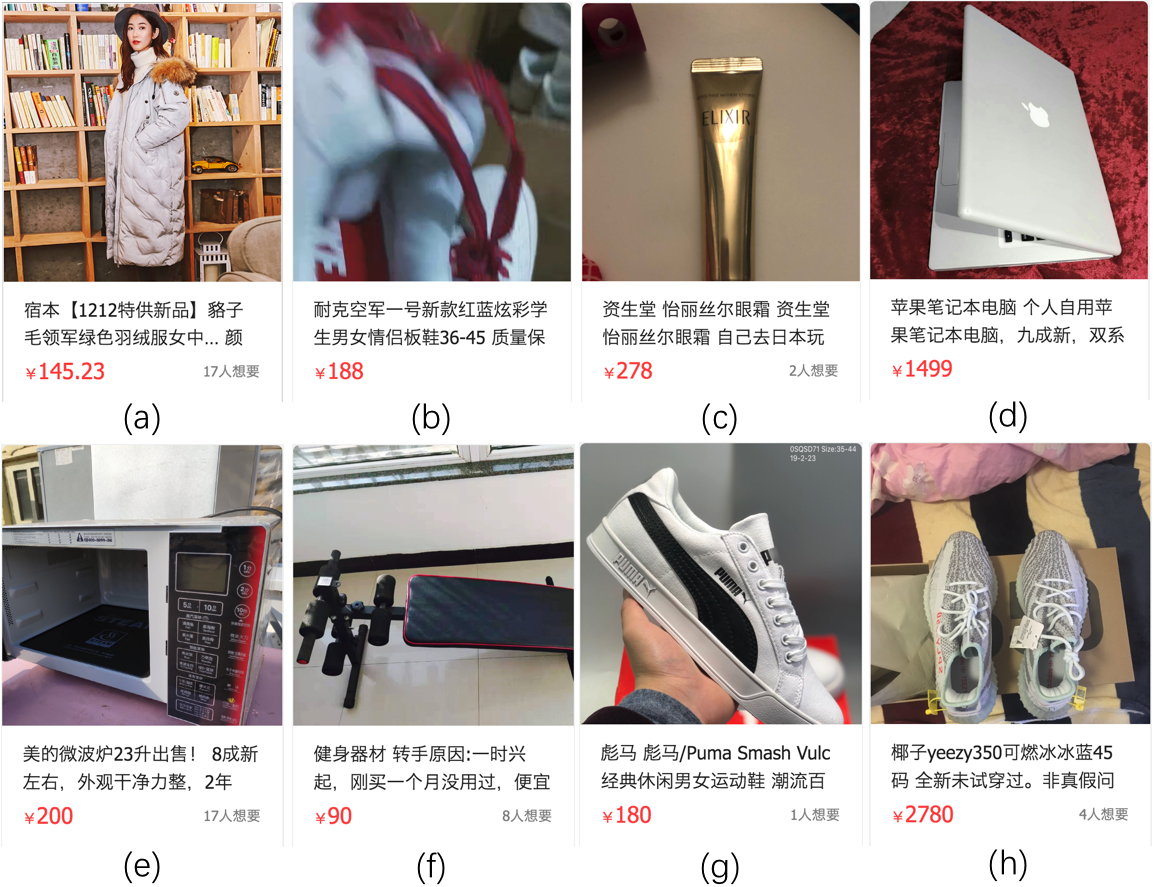}
\caption{Some challenges of vision-based price prediction for online second-hand items. (a) Complex background, (b) blurred image, (c) uneven illumination, (d) inadequate item information, (e,f) rare items, (g,h) brand impact.}
\label{fig:challenge}
\end{figure}

\textbf{Nonstandard image shooting:} In order to predict an item's price with its image, we need to extract representative visual features first. However, unlike the images shot by professional photographers in a controlled photography studio, the second-hand item images are usually taken by the sellers in their daily lives, and thus are with a wide range of varieties. The item in the image may be surrounded by a complex background (Figure~\ref{fig:challenge}(a)), blurred (Figure~\ref{fig:challenge}(b)), or shot with a bad illumination (Figure~\ref{fig:challenge}(c)), which greatly increases the difficulties of precisely extracting the useful features of the item from the image.

\textbf{Inadequate image information:} A picture is worth of a thousand words. However, for E-commerce, an image may not be able to provide the complete information of an item. For the Macbook in Figure~\ref{fig:challenge}(d), even a human expert can hardly predict a reasonable price for this Macbook without any additional specification information, such as the model, storage, memory, etc.

\textbf{Numerous fine-grained categories:} Unlike Airbnb where people mainly share their rooms, or Zipcar that focuses on helping people share their cars, items listed on online second-hand platforms may fall into numerous fine-grained categories, ranging from common categories such as clothes (Figure~\ref{fig:challenge}(a)), cosmetics (Figure~\ref{fig:challenge}(c)), laptops (Figure~\ref{fig:challenge}(d)), to uncommon categories such as household appliances (Figure~\ref{fig:challenge}(e)), fitness equipments (Figure~\ref{fig:challenge}(f)), etc. The price distribution can vary greatly for different categories.

\textbf{Brand impact:} The influence of the item brand should not be overlooked. Figure~\ref{fig:challenge}(g) and Figure~\ref{fig:challenge}(h) show two pairs of shoes with different brands. Though they are both men shoes, the prices of them are quite different. The brand impact is quite common for many kinds of items such as clothes, cosmetics, phones, etc.

In spite of these challenges, we still have some strengths to help us attack this challenging vision-based price prediction problem. First, each listed second-hand item contains an image, from which we can try to extract visual features to perform vision-based price prediction. Second, some other image-related features of the item can be collected to assist the visual feature extraction, such as category, brand, etc. Third, with the increasing prevalence of online second-hand market, we can easily collect tons of data over time to develop the price prediction system and keep refining it.

With these challenges and strengths in mind, we design a vision-based price suggestion system for second-hand items listed on online platforms. First, we use some product information \footnote{In this work, the image-related information we used include attribute, brand, category, and specification. For example, ``a white cotton campus-style adidas T-shirt with size of XL'', attribute: white, cotton; brand: adidas; category: T-shirt; specification: XL, campus-style.} to guide the visual feature extraction from images. With the extracted visual features and some statistical features \footnote{The statistical features used in this work include the first quartile, second quartile, third quartile and mean of the sold price of all historical second-hand items; the first quartile, second quartile, third quartile and mean of the sold price of all historical second-hand items in the same category as the item whose price to be predicted; the first quartile, second quartile, third quartile and mean of the sold price of all historical second-hand items sold by this seller.}, we then jointly train a classification model to determine whether the uploaded image of an item is qualified for price prediction, and a regression model to perform price prediction for items with qualified images. For items with unqualified images, we encourage the sellers to either update the uploaded images (i.e., the current image may have bad qualities such as blur, uneven illumination, etc.) or add some complementary item information such as text description (for those items whose images have inadequate information). Then, we can provide price suggestions for these items with the updated images and/or the complementary text description. The solution for items with unqualified images involves human interaction, which is beyond the scope of this work. In this paper, we focus on the price suggestion using image information only. Our main contributions are summarized as follows:

\begin{figure*}
\includegraphics[width=\linewidth]{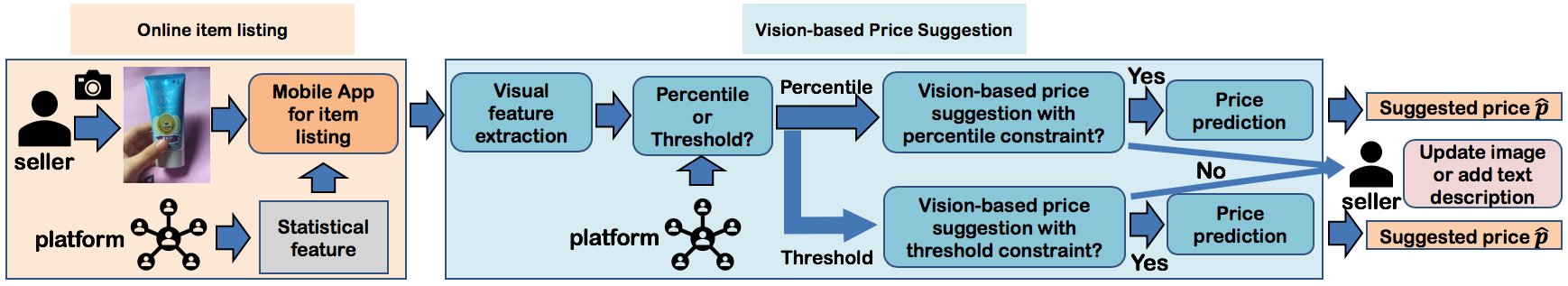}
\caption{Flowchart of the proposed vision-based price suggestion system.}
\label{fig:flowchart}
\end{figure*}

\begin{itemize}
    \item We proposed a vision-based price suggestion system for second-hand items listed on online platforms, which is able to provide accurate or reasonable price suggestions for these second-hand items with their uploaded images.
    
    \item An effective feature extraction method was presented to better extract visual features from the images with the assistance of some other image-related information such as item attribute, brand, category, and specification.
    
    \item A joint optimization framework was designed to simultaneously train (1) a binary classification model to determine whether an uploaded image is qualified for price prediction, and (2) a regression model to provide the price suggestions for items with qualified images. In this module, according to different requirements of image quality for vision-based price prediction (predefined by the platform operator), two specific loss functions were designed for the training of the vision-based price suggestion module which consists of a binary classification model and a regression model.
    
    \item We proposed a training procedure for the joint optimization framework, which can train the classification model and the regression model more stably and with a better performance.

\end{itemize}

\section{Related Work}
\label{sec:rework}

The goal of price prediction for online second-hand items is to suggest a listing price to the seller which will be accepted by both the seller and the potential buyer with a high likelihood. 

A number of previous works focus on developing effective pricing strategies for many different kinds of products. Kusan et al. ~\cite{kucsan2010} use the fuzzy logic theory to develop a grading model for house-building price prediction, which takes the city plans, the nearness to medical, educational and cultural buildings, the public transportation systems and some other environmental factors as the influence factors of the house price. In ~\cite{pudaruth2014}, the author proposes to predict the price of second-hand cars based on some car characters such as model, make, mileage, volume of cylinder, with several different supervised machine learning algorithms. Schlosser and Boissier ~\cite{schlosser2018} propose to derive price reaction strategies with a dynamic programming model in competitive markets with multiple offer dimensions, such as quality, rating and price, which is evaluated by a seller on Amazon for the sale of second-hand books.

Recently, deep neural networks have shown great power on various computer vision tasks such as semantic segmentation ~\cite{pinheiro2015,he2017mask}, object detection ~\cite{cai2018cascade,dai2016}, generative modeling ~\cite{goodfellow2014generative,radford2015unsupervised,dong2018}, and also price prediction ~\cite{peterson2009neural,long2019deep}. In ~\cite{sun2017} an optimized back-propagation neural network algorithm is used to build a price evaluation model, which is based on circulated vehicle data along with a huge number of vehicle transaction data, to evaluate prices of second-hand cars. You et al. ~\cite{you2017} employ a recurrent neural network to estimate the prices of online real estates by taking both the deep visual features extracted from house pictures and the location information into consideration, in which the visual attributes play a major role. Law et al. ~\cite{law2018take} propose to predict house prices with traditional housing features such as size, age and style as well as visual features extracted from both street images and satellite images by a deep neural network model, which is considered to be able to better capture the urban qualities and improve the price estimation. Ye et al. ~\cite{ye2018} present a customized regression model for the dynamic pricing at Airbnb by considering listing features (room type, capacity, locations. etc), temporal information (seasonality), and supply and demand dynamics in the model. Maestre et al. ~\cite{maestre2018} propose to use reinforcement learning to solve a dynamic pricing problem, in which prices are maximized while a balance between the revenue and fairness is well kept. These pricing algorithms work well for a certain item category such as cars or houses, while in our scenarios, we try to provide price suggestions for online second-hand items in numerous fine-grained categories.

Mercari, a Japanese E-commerce company, held a price suggestion challenge \footnote{https://www.kaggle.com/c/mercari-price-suggestion-challenge} recently to provide price suggestions to online sellers by estimating the prices for products they are selling. This challenge does the very similar task as ours. However, no visual feature of the product was involved in their competition.

\section{Vision-based Price Suggestion}
\label{sec:method}

The proposed vision-based price suggestion system aims at providing effective price suggestions for online listed second-hand items based on the visual features extracted from the uploaded images along with some statistical features. Figure~\ref{fig:flowchart} presents the overall price suggestion process. When a seller takes an image for the second-hand item and uploads it to the mobile application for item listing, the uploaded image and some historical statistical features obtained from the platform will be collected by the mobile App. Then, visual features will be extracted from the image (Section~\ref{sec:Feature}). The platform operator can enforce different constraints on second-hand item price suggestion: (1) percentile constraint (i.e., the percent of second-hand items on the platform whose images are qualified for vision-based price suggestion), and (2) threshold constraint (i.e., the second-hand item whose image is qualified for vision-based price suggestion should have its price prediction error less than a given threshold). With these two different operation strategies, two price suggestion modules are designed: (1) vision-based price suggestion with percentile constraint (Section~\ref{sec:percentile}) and (2) vision-based price suggestion with threshold constraint (Section~\ref{sec:threshold}). In each module, a vision-based classification model takes as input the extract visual features along with the statistical features, and determine whether the uploaded image is qualified for price suggestion. If yes, a suggested price of this listed item will be provided to the user by a vision-based regression model, otherwise, the user will be requested to update the item image or provide some text descriptions for better price suggestion (this is not included in this paper).


\begin{figure}
\includegraphics[width=\linewidth]{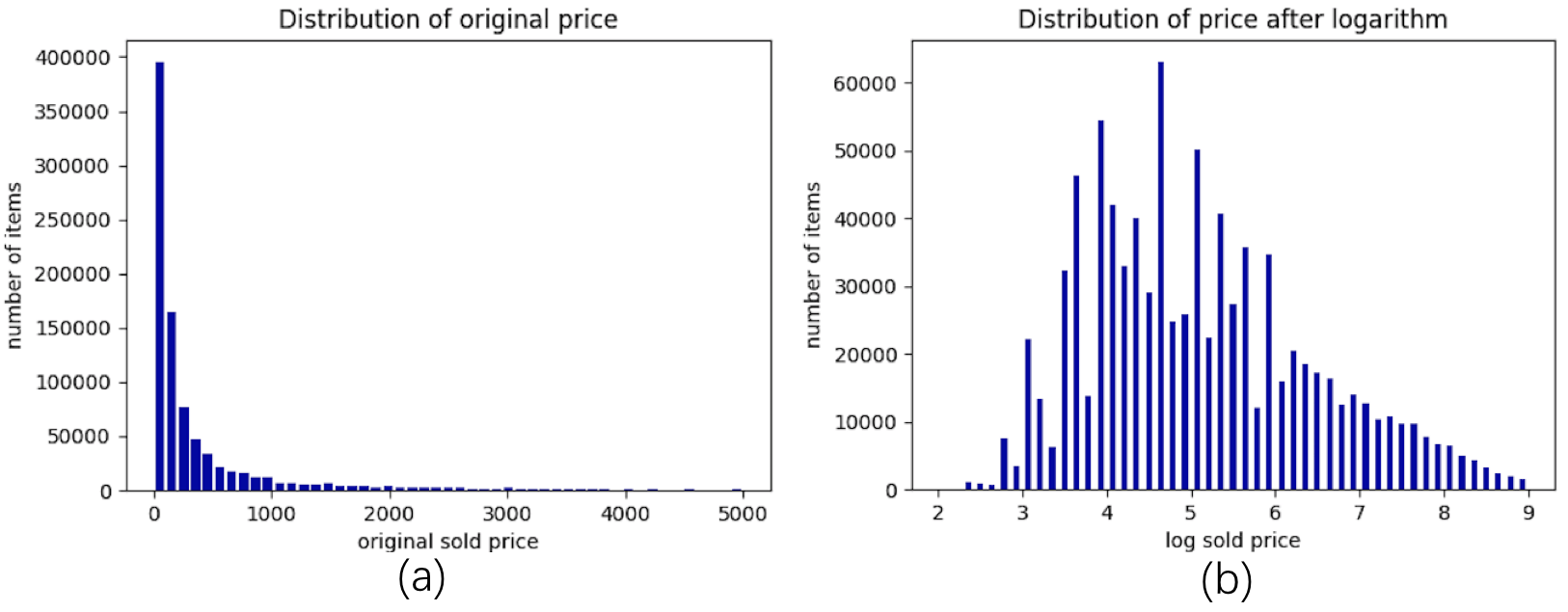}
\caption{Distribution of the original price (a) and the price after the logarithm operation (b).}
\label{fig:price}
\end{figure}

\subsection{Data Preparation and Analysis}
\label{sec:Data}

We collected an online dataset which contains 1,016,445 second-hand items over three months to develop our vision-based price suggestion system. As the dataset accumulated over days contains abnormal data samples, first we delete the items with irrationally low or high prices from the dataset, which removes 5\% of the original data. Then, we check the refined data and find that the price distribution of these online second-hand items is heavily long-tailed, as shown in Figure~\ref{fig:price}(a). It is known that a Gaussian distribution can be better modeled by a model trained with objective functions such as mean square error loss, compared to a long-tailed distribution. Thus, we apply the logarithm operation on the original second-hand item prices, and transform the long-tailed distribution of online second-hand items close to a Gaussian distribution (Figure~\ref{fig:price}(b)).

\subsection{Visual Feature Extraction}
\label{sec:Feature}

Extracting effective features from the image is the most fundamental yet crucial step for developing a vision-based price prediction model. When collecting the item image from the platform, we find that some additional image-related information can be collected as well, such as category, brand, etc. 
Unfortunately, even though each item has its own image, not all of these additional characteristics can be collected for each item in the listing process. Some items may have noisy category label, while some may have missing brand information. Thus, directly taking these image-related characteristics as the input to the prediction model is not a good idea.

\begin{figure}
\includegraphics[width=\linewidth]{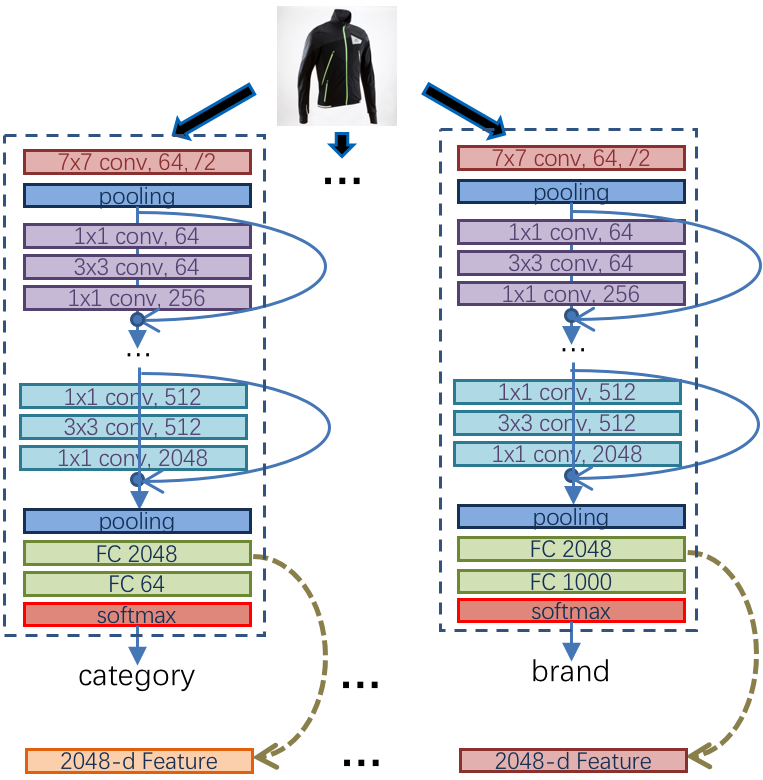}
\caption{Network architecture for visual feature extraction. Multiple single-task models are trained to extract different visual features. ResNet-50 is adopted to build the architecture of each single-task model (denoted by dashed blue box).}
\label{fig:architecture1}
\end{figure}

Inspired by the multi-branch feature extraction model proposed in ~\cite{zhang2018}, which has the ability of simultaneously performing object detection from images with background clutter and representative visual feature extraction of the detected object, we consider to build a visual feature extraction model to extract informative and representative visual features from the item image with its product information (e.g., category, brand, etc.). In this work, instead of building one single multi-task model, we propose to build multiple single-task models to extract visual features. Figure~\ref{fig:architecture1} shows the idea of the proposed visual feature extraction. First, each single-task model takes as input the item image, and performs a specific task (i.e., category prediction, attribute prediction, specification prediction and brand prediction). After each single-task model is well-trained, we employ these single-task models on a new item image, and the feature vectors generated by the second last layer of each model are picked as the extracted visual features of this image. We use ResNet-50 for the architecture building of each single-task model, which is denoted by the dashed blue box in Figure~\ref{fig:architecture1}.

\subsection{Design of Vision-based Price Suggestion}
\label{sec:logic}

\begin{figure}
\includegraphics[width=\linewidth]{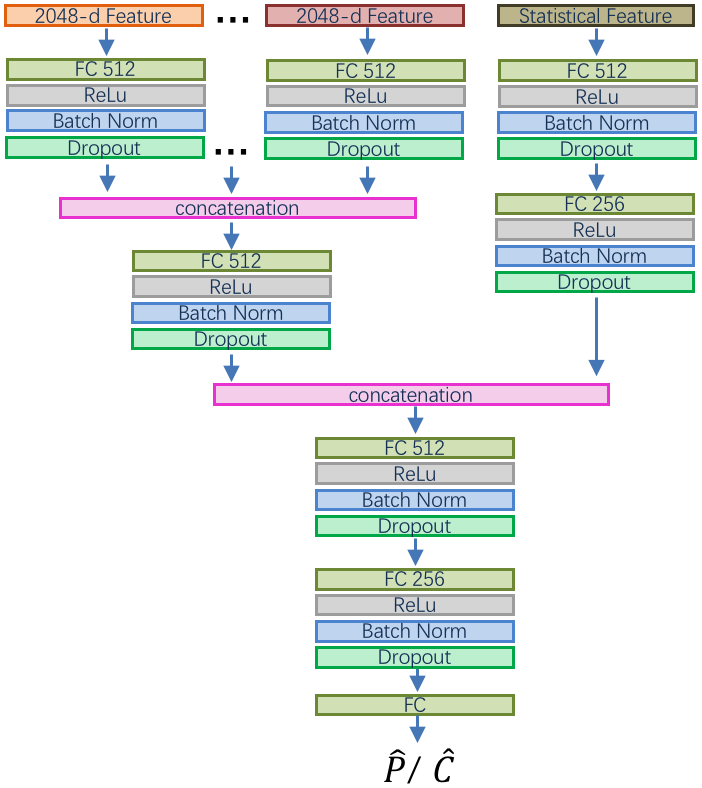}
\caption{Architecture of the regression/classification model. The model takes as input the extracted visual features of a second-hand item along with some statistical features, and provides suggested price for this item (regression model) or determines whether the extracted visual features of the item are qualified for price prediction (classification model).}
\label{fig:architecture2}
\end{figure}

After extracting the visual features, the next step is to train a model with the extracted features to perform price prediction. However, as discussed in the challenges in Section~\ref{sec:intro}, some uploaded images are with inadequate item information or bad image quality (blur, uneven illumination), and accordingly the features extracted from these images are not qualified for offering reasonable price suggestions for the items. Thus, instead of providing a suggested price for a second-hand item without considering the image quality, first we need to determine whether an uploaded image can be used for offering an acceptable price suggestion. In other words, we need a binary classification model to determine whether an item image is qualified for reasonable price prediction, and a regression model to perform price prediction for items with qualified images. We propose to use deep neural networks for these two models, and the same architecture is shared by both the classification model and the regression model except for the output (these two models do not share the weights). Figure~\ref{fig:architecture2} presents the architecture of the regression/classification model, which is a feed-forward neural network. The network takes as input the extracted visual features of the item and some statistical features, and outputs a predicted price for the second-hand item (regression model) or a judgement of whether the extracted visual features of the item are qualified for price prediction (classification model).

Basically, the training of the binary classification model requires some positive samples (images which are qualified for reasonable price suggestions) and negative samples (images which are not qualified for reasonable price suggestions), and the separation of positive and negative samples is usually determined by the price prediction regression model. In turn, the performance of the binary classification model will greatly influence the price prediction model, as it performs price suggestion only on positive items (items with qualified images for reasonable price suggestions). Since the classification and regression tasks are mingled together, instead of training the classification model and regression model separately, we consider to jointly train them.

\subsubsection{Vision-Based Joint Optimization with Percentile Constraint}
\label{sec:percentile}
There are different criteria to determine whether an item image is qualified for offering a reasonable price suggestion. 
From the platform's point of view, the operator can use a percentile strategy, e.g., 50\% of the item images are qualified for price suggestion, and the other 50\% are not. In this scenario, the labels of the samples for training the classification model (i.e., qualified or non-qualified for vision-based price prediction) is not manually determined by human beings, but automatically determined by the regression model. In other word, the classification model regards the items with the top 50\% best price suggestions provided by the regression model as positive, and others as negative. We design a loss function to jointly optimize the classification model and the regression model with the percentile constraint:

\begin{equation}
\begin{split}
    \min_{\Theta_{1},\Theta_{2}} \Big\{&\frac{1}{N}\sum_{i=1}^{N}C_{1}(x_{i};\Theta_{1}) \cdot \Big(y_{i}-f_{reg}(x_{i};\Theta_{2})\Big)^{2} \\ &+ \beta \cdot max\Big(0, \delta-\frac{1}{N}\sum_{i=1}^{N}C_{1}(x_{i};\Theta_{1})\Big) \Big\}
\end{split}
\label{Eq:percent}
\end{equation}
where $C_{1}$ is an indicator function, which is defined as 
\begin{equation}
    C_{1}(x_{i};\Theta_{1}) = \begin{cases}
    0& \text{if $f_{cls}(x_{i};\Theta_{1}) < 0.5$}\\
    1& \text{otherwise},
\end{cases}
\label{Eq:C1}
\end{equation}
where $f_{cls}$ and $f_{reg}$ represent the binary classification model and the regression model, respectively, whose architectures are shown in Figure~\ref{fig:architecture2}; $\Theta_{1}$ and $\Theta_{2}$ are the parameters of $f_{cls}$ and $f_{reg}$, respectively; $N$ is the number of training samples; $x_{i}$ is the $i^{th}$ training sample; $y_{i}$ is the sold price \footnote{The price at which an item is sold. Sold price is usually regarded as the true price of an item because this price is accepted by both the seller and the buyer.} of the training sample $x_{i}$; $\beta$ is a weight to balance these two terms; $\delta$ controls the percentage of positive items (items with qualified images for reasonable price suggestion), i.e., how many items can be regarded as positive items.

The first term in Eq.\ref{Eq:percent} tries to achieve two goals: first, training the price prediction regression model $f_{reg}$ by minimizing a square loss with the positive samples labeled by the binary classification model; and second, training the binary classification model $f_{cls}$ by requiring the classification model to label the samples with the best price suggestions (the differences between the predicted prices given by the regression model and the sold prices are the smallest) as positive (the item image is qualified for reasonable price suggestion). The second term in this equation is a constraint on at least how many items should be regarded as positive items.

\subsubsection{Vision-Based Joint Optimization with Threshold Constraint}
\label{sec:threshold}
The percentile-based joint optimization requires the platform operator to have some prior knowledge of the quality of the item images to set the suitable percentile value. Another criterion for labeling the samples to train the binary classification model is to set a threshold $\epsilon$ for the price prediction accuracy, i.e., if the difference between the predicted price given by the regression model and the sold price of an item is smaller than $\epsilon$, this item is regarded as a positive training sample, otherwise, it is negative. In this scenario, the loss function for the joint optimization of the regression model and the classification model will be
\begin{small}
\begin{equation}
\begin{split}
    \min_{\Theta_{1},\Theta_{2}} \Big\{\frac{1}{N}\sum_{i=1}^{N}C_{1}(&x_{i};\Theta_{1}) \cdot \Big(y_{i}-f_{reg}(x_{i};\Theta_{2})\Big)^{2} \\ 
    + \beta \cdot max\Big(&0, \delta-\frac{1}{N}\sum_{i=1}^{N}C_{1}(x_{i};\Theta_{1})\Big) \\
    + \gamma \!\cdot\! \frac{1}{N}\!\sum_{i=1}^{N}\!\Big[&-C_{2}(x_{i},y_{i};\Theta_{2}) \cdot log\big(f_{cls}(x_{i}; \Theta_{1})\big) \\
    &\!-\! \Big(1\!-\!C_{2}(x_{i},y_{i};\Theta_{2})\Big) \!\cdot\! \Big(log\big(1\!-\!f_{cls}(x_{i}; \Theta_{1})\big)\Big)\Big]
    \Big\}
\end{split}
\label{Eq:threshold}
\end{equation}
\end{small}
where $\gamma$ is a parameter to scale the cross-entropy term, $C_{1}$ is defined as the same as in Eq.\ref{Eq:C1}, and $C_{2}$ is another indicator function which is defined as 
\begin{equation}
    C_{2}(x_{i}, y_{i};\Theta_{2}) = \begin{cases}
    0& \text{if $|y_{i} - f_{reg}(x_{i};\Theta_{2})| > \epsilon$}\\
    1& \text{otherwise}
\end{cases}
\label{Eq:C2}
\end{equation}

The first term in Eq.\ref{Eq:threshold} plays the same role as the first term in Eq.\ref{Eq:percent}. The second term in Eq.\ref{Eq:threshold} makes a constraint to avoid a trivial solution that the binary classification model classifies all the samples as negative, and the regression model offers very bad price suggestions for all the samples. The third term in Eq.\ref{Eq:threshold} is a cross entropy loss to train the binary classification with the training samples labeled by the regression model.

\begin{table*}
\begin{center}
\caption{Summary of dataset used in our experiments.} \label{tab:data}
\resizebox{\textwidth}{!}{
\begin{tabular}{ccccccccccccccc}
  \hline
  class & 1 & 2 & 3 & 4 & 5 & 6 & 7 & 8 & 9 & 10 & 11 & 12 & 13 & total \\
  \hline
  \hline
  \# training sample & 184401 & 38292 & 11196 & 27215 & 16456 & 31362 & 125684 & 65357 & 184861 & 59221 & 88055 & 23330 & 36984 & 892414 \\
  \hline
  \# validation sample & 5528 & 1196 & 290 & 817 & 484 & 972 & 3671 & 1926 & 5360 & 1747 & 2721 & 710 & 1029 & 26451 \\
  \hline
  \# testing sample & 20712 & 4359 & 1055 & 2895 & 1825 & 3656 & 13705 & 7253 & 20060 & 6323 & 9070 & 2503 & 4164 & 97580 \\
  \hline
\end{tabular}}
\end{center}
\end{table*}

\subsection{Warm-up Training}
\label{sec:train}

When jointly training the classification model and the regression model either with Eq.\ref{Eq:percent} or with Eq.\ref{Eq:threshold}, the accuracy of the classification model will be very low at the beginning of the training, and the positive sample determined by the classification model may not exactly be positive, this will fool the training of the regression model, and worsen the performance of the price prediction. In order to obtain a more stable and better training result of joint optimization, we propose to optimize the regression model with all of the training samples at the beginning of the training process. When the classification model is relatively stable, we then use the positive samples labeled by the classification model to further refine the regression model. Specifically, we first let $C_{1}$ in Eq.\ref{Eq:percent} and Eq.\ref{Eq:threshold} equal to 1 for all of the training samples, and jointly optimize the two models with Eq.\ref{Eq:percent} or Eq.\ref{Eq:threshold} (we call this as the first training stage), then we define $C_{1}$ as Eq.\ref{Eq:C1} to continue the training for more iterations (we call this as the second training stage).

In our experiments, we jointly train the classification model and the regression model with a learning rate of 0.0005 for 1,700 epochs, then with a learning rate of 0.0002 for 850 more epochs in the first training stage. In the second training stage, we training these two models with a learning rate of 0.0005 for 3,400 epochs, then with a learning rate of 0.0002 for 1,700 more epochs. We use a batch size of 4096, and the Adam optimizer ~\cite{kingma2014} is adopted for optimization.

\section{Experiments}
We present the experiments in this section. First, we introduce the dataset and evaluation metrics used in our experiments (Section~\ref{sec:data}-~\ref{sec:metric}). Then, we evaluate the vision-based price suggestion both qualitatively and quantitatively (Section~\ref{sec:percenteval}-~\ref{sec:thresholdeval}). Thirdly, we perform ablation studies on the features used in our system (Section~\ref{sec:feature}) and the warm-up training mode (Section~\ref{sec:mode}).

\subsection{Dataset}
\label{sec:data}
We collect a large dataset from Xianyu, an online second-hand platform launched by China's E-commerce giant, Alibaba, to evaluate our proposed system. The data consist of 13 classes: women clothes (1), men clothes (2), children clothes (3), women shoes (4), men shoes (5), children shoes (6), cosmetic (7), phone (8), Computer/Communication/Consumer (3C) product (9), household appliance (10), sports equipment (11), baby product (12), and accessories (13). The data are divided into three sets: training, validation, and testing. Table ~\ref{tab:data} summarizes the data used in our experiments.

\begin{figure*}
\includegraphics[width=\linewidth]{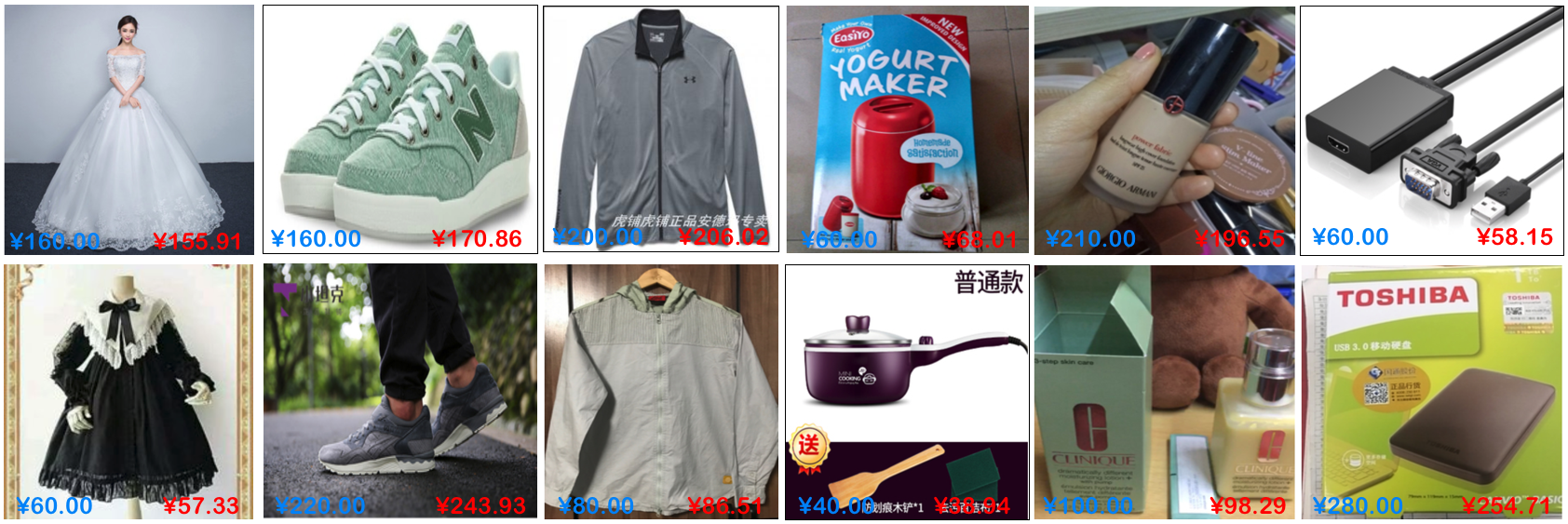}
\caption{Examples of vision-based price suggestion with percentile constraint for positive items. Sold prices are with blue color (left), and predicted prices are with red color (right). All the prices are Chinese Yuan (CHN).}
\label{fig:positive}
\end{figure*}

\begin{figure}
\includegraphics[width=\linewidth]{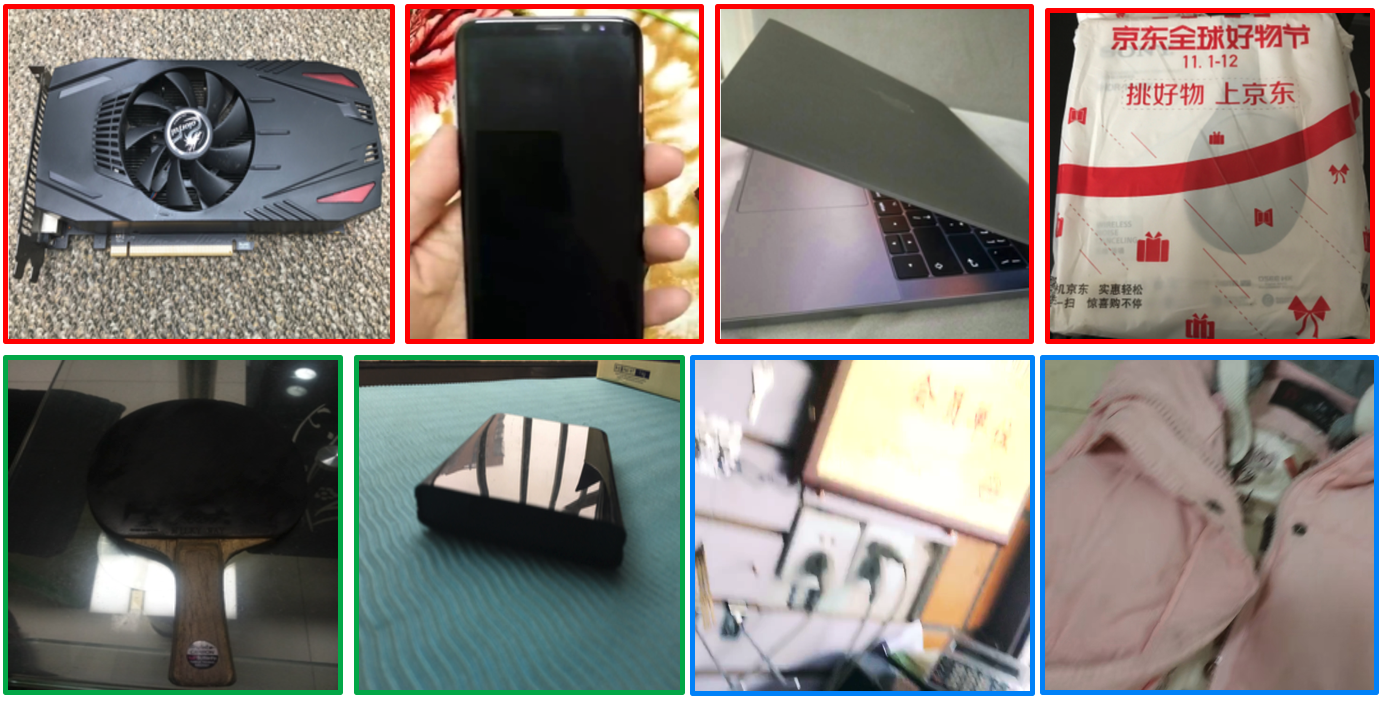}
\caption{Examples of negative items classified by the classification model in the vision-based price suggestion system with percentile constraint. Images with red boundaries have inadequate item information, images with green boundaries are shot in bad illumination, and images with blue boundaries are heavily blurred images.}
\label{fig:negative}
\end{figure}

\subsection{Evaluation Metric}
\label{sec:metric}
The goal of our proposed vision-based price suggestion system is to provide suggested prices for online second-hand items which will be accepted by both the seller and the buyer, so the suggested price should be close to the true price (sold price) of the item. We adopt the Mean Absolute Log Error (MALE) and the Root Mean Square Log Error (RMSLE) to quantitatively evaluate how the suggested prices are close to the true prices of the second-hand items:

\begin{equation}
    MALE = \frac{1}{M}\sum_{i=1}^{M}\Big|log(\hat{p}_{i})-log(p_{i})\Big|
\end{equation}

\begin{equation}
    RMSLE = \sqrt{\frac{1}{M}\sum_{i=1}^{M}\Big(log(\hat{p}_{i})-log(p_{i})\Big)^{2}}
\end{equation}
where $M$ is the number of testing samples, $\hat{p}_{i}$ and $p_{i}$ are the predicted price and sold price of product $i$, respectively.

\begin{figure*}
\includegraphics[width=\linewidth]{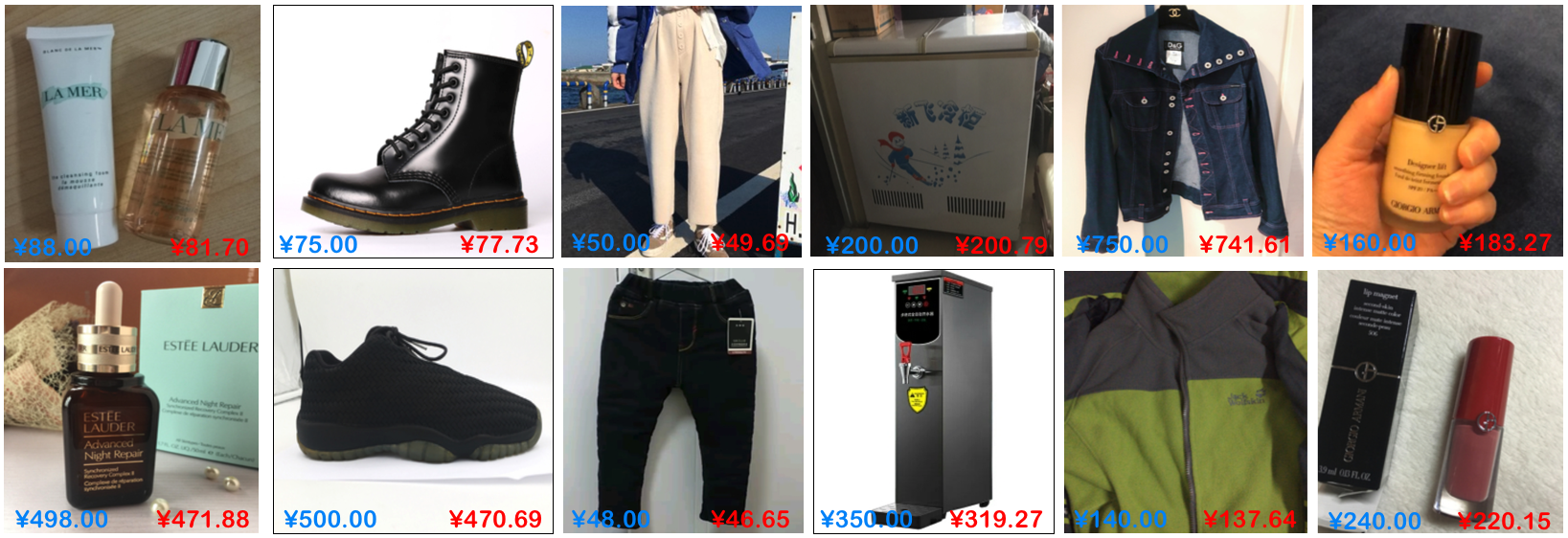}
\caption{Examples of vision-based price suggestion with threshold constraint for positive items. Sold prices are with blue color (left), and predicted prices are with red color (right). All the prices are Chinese Yuan (CHN).}
\label{fig:positive1}
\end{figure*}

\begin{figure}
\includegraphics[width=\linewidth]{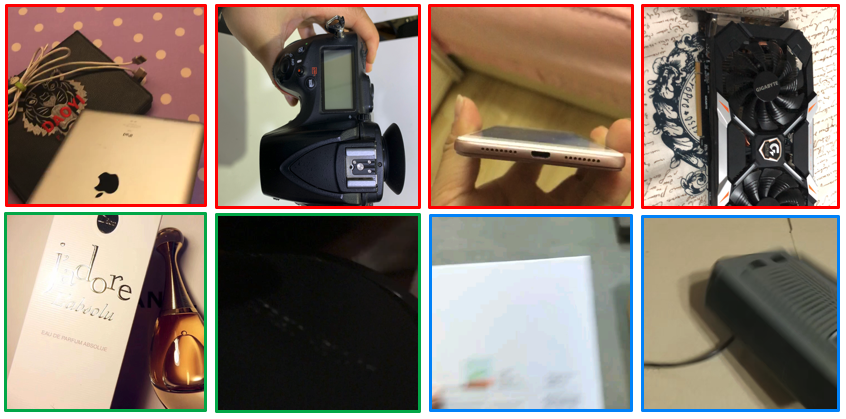}
\caption{Examples of negative items classified by the classification model in the vision-based price suggestion system with threshold constraint. Images with red boundaries have inadequate item information, images with green boundaries are shot in bad illumination, and images with blue boundaries are heavily blurred images.}
\label{fig:negative1}
\end{figure}

\begin{table}
  \caption{Accuracy of vision-based price suggestion with percentile constraint. Positive items are with qualified images for vision-based price suggestion, and the MALE and RMSLE values of the positive items are calculated.}
  \label{tab:percent}
  \resizebox{\columnwidth}{!}{
  \begin{tabular}{|c|c|c|c|c|c|}
    \toprule
    percentile & 30 & 40 & 50 & 60 & 70 \\
    \midrule
    \# positive items & 29468 & 39694 & 48871 & 60781 & 69794\\
    \% positive items & 30.20\% & 40.68\% & 50.08\% & 62.29\% & 71.52\%\\
    MALE & 0.3573 & 0.3685 & 0.3864 & 0.4012 & 0.4132\\
    RMSLE & 0.4828 & 0.5023 & 0.5302 & 0.5502 & 0.5666\\
    \bottomrule
\end{tabular}}
\end{table}

\subsection{Evaluation of Vision-based Price Suggestion with Percentile Constraint}
\label{sec:percenteval}
We first test the performance of the vision-based price suggestion with percentile constraint. In the training stage, the binary classification model ($f_{cls}$ in Eq.~\ref{Eq:percent}) is trained with the percentile constraint given by the platform operator (i.e., how many percents of item samples whose images should be regarded as qualified for vision-based price suggestion). When testing, the classification model classifies the positive items (items whose images are qualified for vision-based price suggestion) from the negative ones (items with unqualified images for vision-based price suggestion), then the regression model ($f_{reg}$ in Eq.~\ref{Eq:percent}) provides suggested prices for the positive items. Table~\ref{tab:percent} summarizes the results, which shows that the percent of positive items classified by the binary classification model is almost exactly the same with the percentile constraint. This demonstrates the effectiveness of the classification model. Besides, as the number of positive items increases (i.e., more items are classified as positive), the $MALE$ and $RMSLE$ values of the positive items increase. This is intuitive, since the quality control for classifying positive item images decreases with the increase of percent of positive items.

Some examples of the vision-based price suggestion with percentile constraint (percentile constraint is 50\%) on positive items are presented in Figure~\ref{fig:positive}. For each second-hand item, the gap between the predicted price and the sold price is very small, which demonstrates the effectiveness of the regression model. Figure~\ref{fig:negative} shows the images of some negative samples classified by the classification model under the percentile restriction of 50\%. Some of the negative samples have images which do not contain sufficient item information (e.g., images with red boundaries in Figure~\ref{fig:negative}), and some are with poor image qualities such as bad illumination (e.g., images with green boundaries in Figure~\ref{fig:negative}) or image blur (e.g., images with blue boundaries in Figure~\ref{fig:negative}) so that the extracted visual features are not representative. Figure~\ref{fig:positive} and Figure~\ref{fig:negative} also illustrate that the classification model is able to effectively classify the positive items and the negative items.

\subsection{Evaluation of Vision-based Price Suggestion with Threshold Constraint}
\label{sec:thresholdeval}
In this subsection, we test the vision-based price suggestion with threshold constraint. Given a threshold value, if the absolute log error of the predicted price and the sold price of a second-hand item is smaller than this threshold, the item is regarded as a positive training sample, otherwise, the item is negative. Accordingly, the joint training of the binary classification model and the regression model is greatly influenced by the given threshold value. Table~\ref{tab:threshold} summaries the performance of the regression model under different threshold constraints.

\begin{table}
  \caption{Quantitative evaluation of the regression model for price prediction under different threshold constraints.}
  \label{tab:threshold}
  \resizebox{\columnwidth}{!}{
  \begin{tabular}{|c|c|c|c|c|c|}
    \toprule
    value of $\epsilon$ in Eq.~\ref{Eq:threshold} & 0.3 & 0.4 & 0.5 & 0.6 & 0.7 \\
    \midrule
    \# positive items & 34256 & 52561 & 67365 & 77264 & 80297 \\
    \% positive items & 35.11\% & 53.86\% & 69.04\% & 79.18\% & 82.29\% \\
    MALE & 0.3427 & 0.3776 & 0.4033 & 0.4265 & 0.4309 \\
    RMSLE & 0.4764 & 0.5223 & 0.5553 & 0.5878 & 0.5917 \\
    \bottomrule
\end{tabular}}
\end{table}

When the given threshold value $\epsilon$ in Eq.~\ref{Eq:threshold} increases, i.e., the online platform looses its restriction on the definition of positive items, more second-hand items are considered to be with qualified images for vision-based price prediction by the binary classification model, thus the regression model provides suggested prices for more second-hand items. In this scenario, it is no doubt that the $MALE$ and $RMSLE$ values of the positive items (classified by the classification model) will increase. We can imagine that if the classification model picks an item with a very high-quality image as a positive item, and the regression model offers suggested price only for this positive item, the $MALE$ and $RMSLE$ values will be very low, but this is not sufficient to demonstrate that the performance of the classification model and the regression model are good. Thus, when evaluating these two models, we should not only pay attention to the $MALE$ and $RMSLE$ metrics of the positive items, but also keep our eyes on the number of positive items.

Some positive items and negative items classified by the binary classification model in the vision-based price suggestion module with threshold constraint (threshold value $\epsilon$=0.5) are presented in Figure~\ref{fig:positive1} and Figure~\ref{fig:negative1}, with the predicted price by the regression model in this price suggestion system listed on the bottom right of each item image. These two figures also demonstrate the effectiveness of the classification model and the regression model in the vision-based price suggestion module with threshold constraint.

\subsection{Evaluation of Feature Extraction}
\label{sec:feature}
To evaluate the effectiveness of the proposed visual feature extraction strategy, we perform two comparative experiments. In one experiment, we adopt the proposed visual feature extraction strategy (i.e., extract visual features with the assistance of image-related characteristics, we call it multiple single-task feature extraction (MSTFE)) to train the vision-based price prediction module with percentile constraint (percent constraint on positive items is 50\%), and then we train this module again without using the proposed visual feature extraction strategy (i.e., a ResNet-50 architecture is employed to extract visual features from the image of an item directly to do the price prediction without using the image-related characteristics, we call it single-task feature extraction (STFE)). We also train the vision-based price prediction module with threshold constraint (threshold value $\epsilon$ in Eq.~\ref{Eq:threshold} is 0.5) with and without the proposed feature extraction strategy, respectively. The experiment results are presented in Table~\ref{tab:feature}, which shows that the proposed visual feature extraction strategy is more effective on extracting representative visual features for vision-based price prediction.

\begin{table}
  \caption{Quantitative evaluation of proposed visual feature extraction strategy.}
  \label{tab:feature}
  \resizebox{\columnwidth}{!}{
  \begin{tabular}{|c|c|c|c|c|}
    \hline
    constraint & feature & \# positive items & MALE & RMSLE \\
    \hline
    \hline
    \multirow{2}{*}{percentile} & STFE & 46583 & 0.4008 & 0.5421 \\
    \cline{2-5}
               ~ & MSTFE & 48871 & 0.3864 & 0.5302 \\
    \hline
    \hline
    \multirow{2}{*}{threshold} & STFE & 58949 & 0.4104 & 0.5572 \\
    \cline{2-5}
               ~ & MSTFE & 67365 & 0.4033 & 0.5553 \\
    \hline
\end{tabular}}
\end{table}

Since we use the extracted feature vector, not the predicted brand, category, attribute and specification, as the input of the regression model for price prediction, given the images of two iPhone4, the model will not output an average price of all iPhone4, but predicts different prices for these two iPhone4 according to the different extracted visual features of these two items.

\subsection{Evaluation of the Training Mode}
\label{sec:mode}
Finally, we evaluate the effectiveness of the proposed warm-up training mode. Two comparative experiments are designed. One is training the vision-based price suggestion module with percentile constraint (percent constraint on positive items is 50\%) with and without warm-up training mode, respectively, and the other is training the vision-based price suggestion module with threshold constraint (threshold value $\epsilon$ in Eq.~\ref{Eq:threshold} is 0.5) with and without warm-up training mode, respectively. The results are summarized in Table~\ref{tab:mode}, from which we can see that the models trained with the warm-up training are with better performance.

\begin{table}
  \caption{Evaluation of proposed warm-up training mode.}
  \label{tab:mode}
  \resizebox{\columnwidth}{!}{
  \begin{tabular}{|c|c|c|c|c|}
    \hline
    constraint & training mode & \# positive items & MALE & RMSLE \\
    \hline
    \hline
    \multirow{2}{*}{percentile} & W/O warm-up & 48130 & 0.3940 & 0.5343 \\
    \cline{2-5}
               ~ & W warm-up & 48871 & 0.3864 & 0.5302 \\
    \hline
    \hline
    \multirow{2}{*}{threshold} & W/O warm-up & 55575 & 0.4058 & 0.5484 \\
    \cline{2-5}
               ~ & W warm-up & 67365 & 0.4033 & 0.5553 \\
    \hline
\end{tabular}}
\end{table}

\section{Conclusion}
\label{sec:conclusion}
In this work, we presented a vision-based price prediction system to provide price suggestions for online second-hand items with the uploaded item images, which can help the second-hand item sellers set listing prices for the item effectively. In the proposed system, we design different loss functions to train the price suggestion system, so that various demands from the platform operator can be satisfied. Experiment results on a large real-world dataset demonstrate the effectiveness of the proposed price suggestion system.


{\small

}

\end{document}